\documentclass[twocol]{ametsocV6.1}

\usepackage{soul}
\usepackage{float}

\usepackage{algorithm}
\usepackage{algpseudocode}

\title{EC-EarthFlow: Probabilistic emulation of daily transient global climate model simulations with flow matching}

\authors{Kirien Whan$^*$,\thanks{$^*$These authors contributed equally to this work.}\aff{a}\correspondingauthor{Kirien Whan, kirien.whan@knmi.nl} 
Nikolaj T. M{\"u}cke$^*$,\aff{b, c}\footnotemark[\value{footnote}] 
and Karin van der Wiel,\aff{a, d} 
}

\affiliation{\aff{a}{KNMI, Utrechtseweg 297, De Bilt, 3731GA , The Netherlands}\\
\aff{b}{Scientific Computing, Centrum Wiskunde \& Informatica, Science Park 123, Amsterdam, 1098XG, The Netherlands}\\
\aff{c}{Faculty of Civil Engineering and Geosciences, Delft University of Technology, Delft, The Netherlands}\\
\aff{d}{Faculty of Geosciences, Utrecht University, Utrecht, The Netherlands}
}

\abstract{We introduce \textit{`EC-EarthFlow'}, a generative flow matching model that emulates simulations from the physical climate model EC-Earth3. The model is trained on transient simulations from EC-Earth3 (1950-2166, SSP2-4.5) to predict the day ahead temperature field from the previous days temperature as well as annual mean temperature. Predictions are made auto-regressively with rollout periods of between a month and an extended season. Using only this variable of interest, we are able to reproduce the daily variability, spatial patterns, annual cycle and long-term trend from EC-Earth3 at a substantially lower computational cost than the physical model. We demonstrate that EC-EarthFlow is stable for long inference periods, and that it can learn the physical relationships as simulated in EC-Earth3.} 

\begin{document}

\maketitle

\section{Introduction}

Understanding climate change is highly relevant given the expected impacts that it will have on society in the years and decades ahead \citep{masson2021climate,IPCCWG2}. Global Climate Models (GCMs) are an essential tool for understanding these future changes. GCMs are physics-based models of the Earth system, similar to weather forecasting models. They produce simulations of Earth's climate with realistic spatial, temporal and inter-variable dependence structures, and can be integrated many years in to the future \citep{o2016scenario} to study changes in mean climate, climate variability and (multi-variate or compound) extremes \citep[e.g.][]{tebaldi2021climate,wiel2021contribution,zscheischler2020typology,mourik2026multi}. For the study of climate variability and extremes, often many years of simulated data are required. A key approach is therefore to make use of `large ensembles', many realizations of the same future climate experiment that start from different initial conditions \citep{deser2020insights}. These large ensembles explicitly account for internal variability of the climate system \citep{wiel2019added, muntjewerf2023knmi}, and have become an important tool that climate scientists can use to separate forced responses from internal variability and assess forced changes in variability itself \citep{muntjewerf2023knmi}. However, large-ensembles are costly to run, and so they are a prime candidate for the use deep learning, and specifically for the use of generative AI methods where the productions of ensembles is straight-forward \citep{tomczak2024deep}.

The use of deep learning \citep{lang2024aifs, wijnands2025comparison} and generative or ensemble methods \citep{price2025probabilistic, lang2026aifs} is now common practice in weather forecasting but there is less work so far in a climate context. This is likely due to the additional challenges that the climate context presents, namely, stability over long time periods, and the non-stationarity of training data and desired simulation data. Auto-regressive deterministic models have been shown to simulate the whole atmosphere well and to be stable over long rollout periods when trained in a stationary climate \citep{watt2023ace, watt2025ace2, chapman2025camulator}. There are several drawbacks to these models. First, they lack a sophisticated probabilistic framework as ensembles of simulations are generated by perturbing the initial states that are used as inputs to the emulator. Second, they have not been used in a non-stationary climate (to our knowledge). Finally, deterministic models are generally known in a weather context for producing smooth forecasts that lack high-frequency variability \citep{lang2026aifs}. Another approach is to use generative methods that learn the conditional distributions \citep[e.g.][]{ruhling2024probablistic, bassetti2024diffesm, brenowitz2025climate}. \citet{ruhling2024probablistic} use a generative model in a stationary climate. \citet{bassetti2024diffesm} train a diffusion model to predict a fixed 30-day sequence of either temperature or precipitation in the current and future climate. They focus on variables of interest (i.e., near-surface temperature and precipitation), likely due to the computational constraints of the more expensive generative model. They show that extremes are well represented in future climates \citep{bassetti2024diffesm}, although one drawback of such a method is the discontinuities that are introduced by the fixed length sequence. `cBottle' is a generative model that samples directly from the full distribution of the full atmospheric states from the period 1980-2024 (with the forced trend included) at very high spatial resolution and does not require a previous time step or a auto-regressive rollout \citep{brenowitz2025climate}. For example, independent hourly samples from 1940-2021 were generated by conditioning on monthly sea-surface temperatures in a model containing calendar embeddings (e.g., day of the year, and time of the day) as well as a positional embedding (e.g., a grid point). The model simulates a trend during this period, although it is weaker than observed. One draw back of this model is that it can't simulate multi-day events, like heatwaves. The video extension is fixed to a 3-day period. This literature demonstrates that generative models are able to simulate the climate system, but that there isn't yet an approach that can emulate daily, transient future climate projections.

In line with the need for (computationally cheaper) large ensembles of transient climate model simulations, our goal is to emulate EC-Earth3, a state-of-the-art GCM \citep{doscher2021ec}. To our knowledge, we are the first to emulate the transient run of a GCM, at the daily scale, including the forced trend over the period 1950-2166 and make predictions auto-regressively for arbitrarily long periods. We use a flow matching generative model \cite{lipman2022flow, liu2022flow}. In this first proof-of-concept, we focus on emulating a single high-impact variable, near-surface temperature, with a substantial trend \citep{rahmstorf2017global,brogno2025review}. The focus on near-surface temperature for this first proof-of-concept is for both pragmatic computational and scientific reasons. For the latter, it is of interest to see how well a deep learning model can simulate a single variable with a substantial trend, compared to the full atmosphere usually used in other studies \citep[e.g.]{watt2023ace, brenowitz2025climate}. We discuss the data, methods and implementation in Section~\ref{sec:datamethods}, followed by results and conclusions in Sections~\ref{sec:results} and \ref{sec:conc}.

\section{Data and Methods}\label{sec:datamethods}
In this section we will outline the GCM that we are aiming to emulate, EC-Earth3, as well as the generative models that we use, and the specific details of the implementation.

\subsection{Data} 
The climate model that we emulate in this work is EC-Earth3 \citep{doscher2021ec}. It is a fully-coupled climate model (i.e. simulating atmosphere, ocean, land surface and cryosphere) with a native resolution of T255L91 in the atmosphere (horizontally $\approx 80~km$, corresponding to 512 $\times$ 256 grid cells, and 91 vertical layers). To reduce the computational burden, we decreased the horizontal resolution using block averaging to 128 $\times$ 64 grid cells. We use physics version 5 (EC-Earth3\_p5), a slightly re-tuned model version with comparatively smaller temperature biases in the Northern Hemisphere and a larger Southern Ocean warm bias than EC-Earth3, full details in \citep{muntjewerf2023knmi} and \citep{dorland2023knmi}. We use  data from a SSP2-4.5 experiment, running from 1950 to 2166 \citep{dorland2023knmi}. We discard simulation data before 1950 due to unrealistic oceanic oscillations in the model during this time \citep{dorland2023knmi}. We take daily 2m-temperature from 16 transient runs from the EC-Earth3 ensemble. In the CMOR-compliant output this is the variable '\emph{tas}', in the remainder of this paper we use $T$. In addition to the daily 2m-temperature, we used the annual-mean of 2m-temperature by taking the average at each grid cell of all days in a calendar year ($\bar{T}_{\text{annual}}$).

\subsection{Methods} 
\label{sec:methods}
\subsubsection{Flow Matching Model}
\label{sec:flow_matching}

\begin{figure*}
    \centering
    \includegraphics[width=0.9\linewidth]{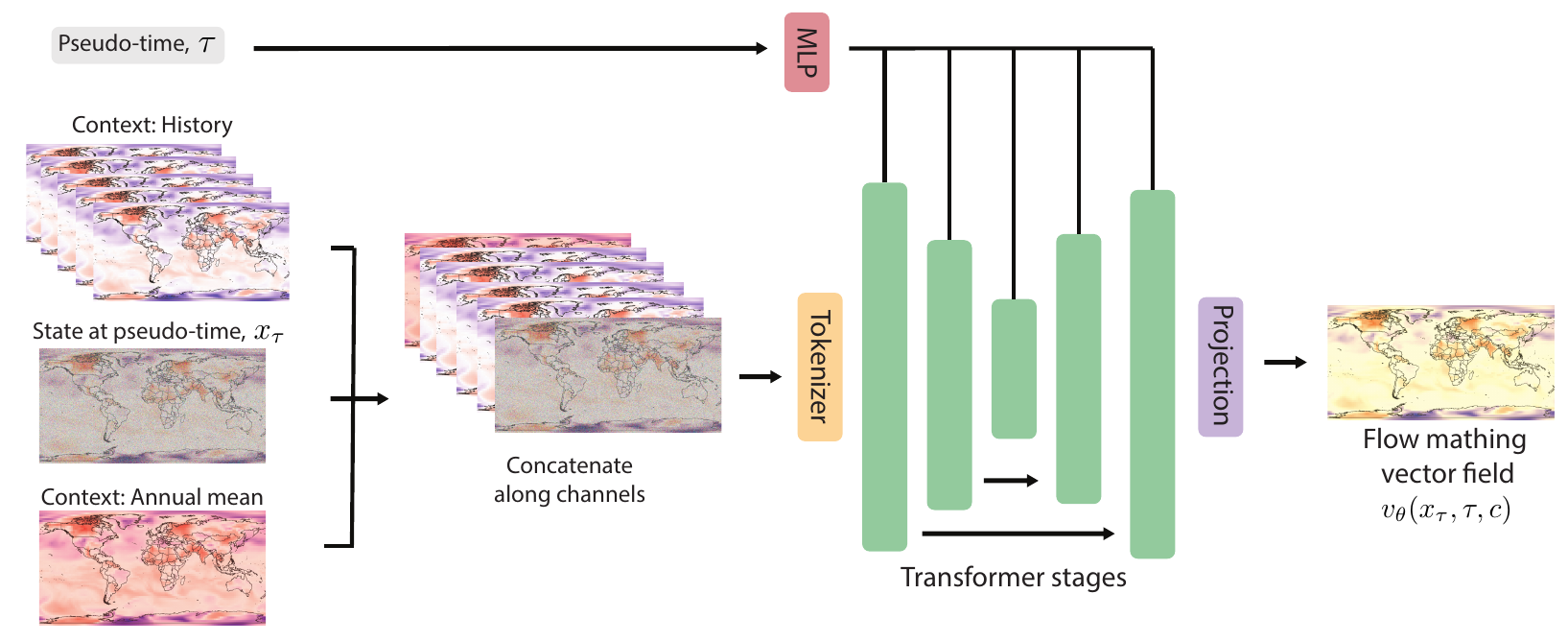}
    \caption{The architecture of EC-EarthFlow showing that pseudo-time is concatenated with contextual inputs and fed into the neural network model to predict the flow matching vector field.}
    \label{fig:architecture}
\end{figure*}

\begin{figure*}
    \centering
    \includegraphics[width=0.9\linewidth]{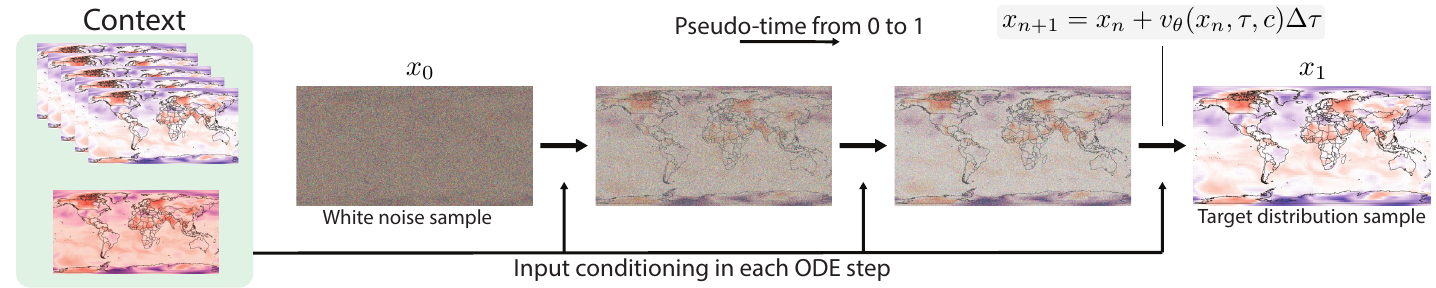}
    \caption{An illustration of the generative flow matching process. The process transforms noise to the target distribution by integrating an ODE over pseudo-time.}
    \label{fig:fm-generation}
\end{figure*}

Our objective is to model the conditional distribution of the 2m-temperature field at the next daily time step, $T_{t+1}$, given a context $c$. The context comprises the previous five days of daily gridded temperature, which carry the recent temporal dynamics, and the gridded annual mean temperature, which fixes the mean climate state: $c = (T_{t-4}, T_{t-3}, T_{t-2}, T_{t-1}, T_{t}, \bar{T}_{\text{annual}})$. The model therefore samples from
\begin{align}
    p(T_{t+1} | c) =  p(T_{t+1} | T_{t-4}, T_{t-3}, T_{t-2}, T_{t-1}, T_{t}, \bar{T}_{\text{annual}})
\end{align}
Flow matching samples the target distribution by transporting a sample $x_0$ from a tractable base distribution into a sample $x_1 = T_{t+1}$ from the data distribution. The transport is defined by an ordinary differential equation in pseudo-time,
\begin{equation}
    \frac{\mathrm{d} x_\tau}{\mathrm{d} \tau} = v_\theta(x_\tau, \tau, c), \quad \tau \in [0, 1],
\end{equation}
where $\tau$ is the artificial integration time (distinct from the physical time $t$), $x_\tau$ the state at time $\tau$, and $v_\theta$ a time-dependent vector field parameterised by a neural network with weights $\theta$ (Figure \ref{fig:architecture}). Integrating from $x_0 \sim \mathcal{N}(0, \mathbf{I})$ yields $x_1 \sim p(\cdot \mid c)$ (Figure \ref{fig:fm-generation}).

We train $v_\theta$ with Conditional Flow Matching (CFM) along an Optimal Transport (OT) probability path \citep{lipman2022flow}, which interpolates linearly between the noise sample $x_0$ and the target sample $x_1$:
\begin{equation}
    x_\tau = (1 - \tau) x_0 + \tau x_1.
\end{equation}
The target vector field along this path is the constant displacement
\begin{equation}
    u_\tau(x_\tau \mid x_0, x_1) = x_1 - x_0.
\end{equation}
On the regular latitude-longitude grid of EC-Earth3, the physical area of a grid cell shrinks toward the poles, so an unweighted loss over-represents polar latitudes. We therefore fit $v_\theta$ to $u_\tau$ using a latitude-weighted mean squared error:
\begin{equation}
    \mathcal{L}(\theta) = \mathbb{E}_{\tau, x_0, x_1, c} \left[ \left\| \mathbf{W}(\phi) \odot \left( v_\theta(x_\tau, \tau, c) - (x_1 - x_0) \right) \right\|_2^2 \right],
\end{equation}
with $\tau \sim \mathcal{U}(0,1)$, $x_0 \sim \mathcal{N}(0, \mathbf{I})$, $x_1 \sim p_{\text{data}}$, $\mathbf{W}(\phi)$ a diagonal weighting matrix with entries proportional to $\cos\phi$, and $\odot$ the element-wise product. All inputs are globally normalized before being passed to the network. Algorithm \ref{alg:flow_matching_train} gives pseudo-code for a single training iteration.

\begin{figure*}
    \centering
    \includegraphics[width=0.9\linewidth]{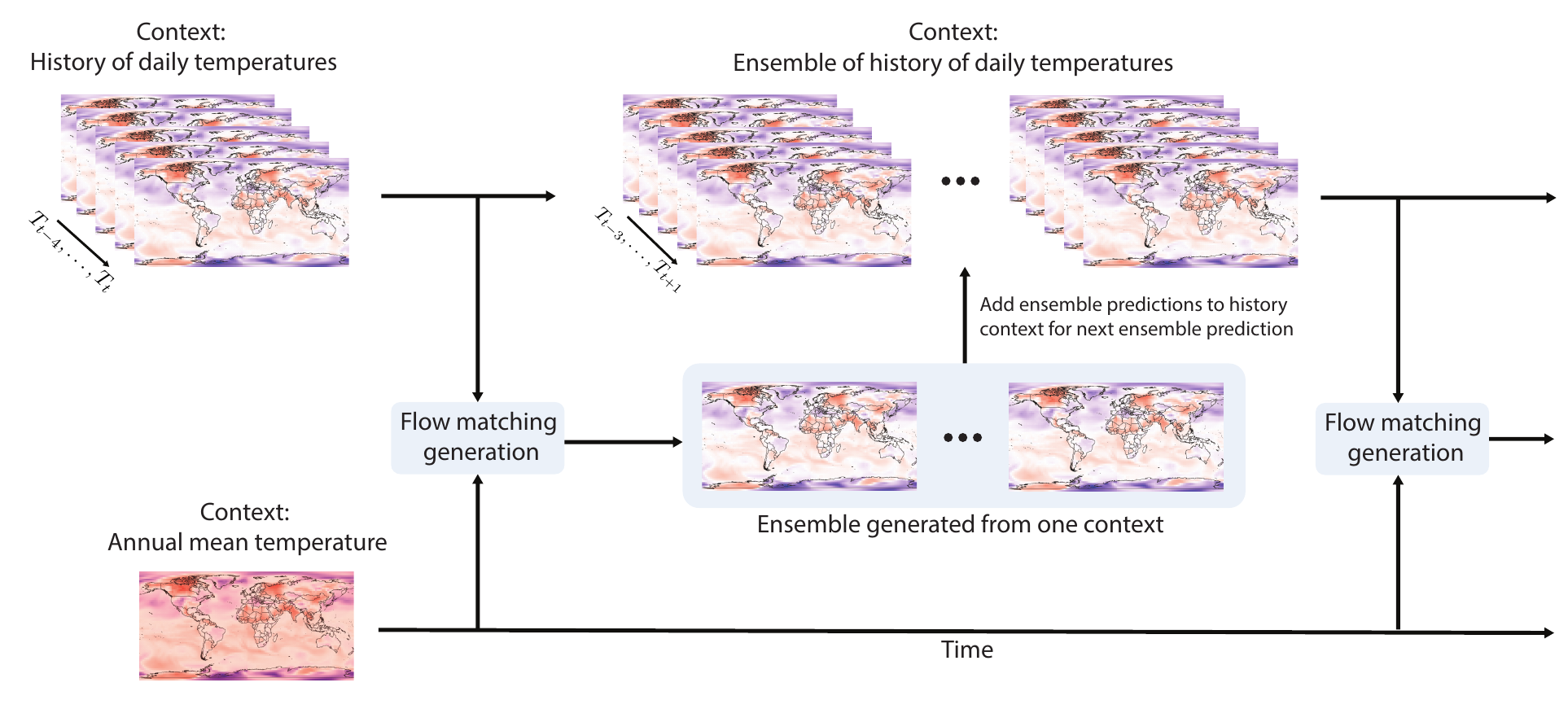}
    \caption{The auto-regressive rollout used at inference. The context information is combined with noise to produce a prediction for the next day. These predictions are then added to the context window for the following prediction. We produce an ensemble of predictions by using different noise samples.}
    \label{fig:fm-inference}
\end{figure*}

At inference, the day-ahead field $T_{t+1}$ is generated by drawing a latent noise vector $x_0$ and integrating the ODE forward to $\tau=1$ conditioned on $c$, using a standard numerical solver. Ensembles follow from $N$ independent noise realizations $\{x_0^{(i)}\}_{i=1}^N$ integrated in parallel. Appending the terminal state to the context yields the conditioning for the next step, giving an autoregressive rollout over longer horizons. Ensemble spread is inherited from the internal variability of the EC-Earth3 training data. See Algorithm \ref{alg:flow_matching_inference} and Figure \ref{fig:fm-inference}.

\begin{algorithm}[H]
\caption{EC-EarthFlow Training Iteration}
\label{alg:flow_matching_train}
\begin{algorithmic}[1]
\Require Training dataset $\mathcal{D}$, neural network $v_\theta$, learning rate $\eta$, latitude weights $\mathbf{W}(\phi)$
\State Sample a physical target and its context:
\Statex \hspace*{\algorithmicindent}$(x_1, c) \sim \mathcal{D}$
\State Sample Gaussian noise: $x_0 \sim \mathcal{N}(0, \mathbf{I})$
\State Sample integration time: $\tau \sim \mathcal{U}(0, 1)$
\State Construct the intermediate state:
\Statex \hspace*{\algorithmicindent}$x_\tau \gets (1 - \tau) x_0 + \tau x_1$
\State Compute the OT target vector field:
\Statex \hspace*{\algorithmicindent}$u_\tau \gets x_1 - x_0$
\State Predict the vector field:
\Statex \hspace*{\algorithmicindent}$\hat{v}_\tau \gets v_\theta(x_\tau, \tau, c)$
\State Compute latitude-weighted loss:
\Statex \hspace*{\algorithmicindent}$\mathcal{L} \gets \left\| \mathbf{W}(\phi) \odot (\hat{v}_\tau - u_\tau) \right\|_2^2$
\State Update network parameters:
\Statex \hspace*{\algorithmicindent}$\theta \gets \theta - \eta \nabla_\theta \mathcal{L}$
\end{algorithmic}
\end{algorithm}

\begin{algorithm}[H]
\caption{EC-EarthFlow Autoregressive Inference}
\label{alg:flow_matching_inference}
\begin{algorithmic}[1]
\Require Trained network $v_\theta$, initial context $c_0$, forecast horizon $H$, ensemble size $N$
\State Initialize ensemble forecast set $\hat{\mathcal{T}} \gets \emptyset$
\For{$i = 1, \dots, N$} \Comment{In parallel}
    \State Initialize context $c \gets c_0$
    \Statex \hspace*{\algorithmicindent}\hspace*{\algorithmicindent}and trajectory $\mathcal{T}^{(i)} \gets \emptyset$
    \For{$t = 1, \dots, H$}
        \State Sample latent noise: $x_0 \sim \mathcal{N}(0, \mathbf{I})$
        \State Integrate ODE to $\tau=1$:
        \Statex \hspace*{\algorithmicindent}\hspace*{\algorithmicindent}\hspace*{\algorithmicindent}$T_t \gets \text{ODESolve}(v_\theta, x_0, \tau \in [0,1], c)$
        \State Append prediction to trajectory:
        \Statex \hspace*{\algorithmicindent}\hspace*{\algorithmicindent}\hspace*{\algorithmicindent}$\mathcal{T}^{(i)} \gets \mathcal{T}^{(i)} \cup \{T_t\}$
        \State Update context auto-regressively:
        \Statex \hspace*{\algorithmicindent}\hspace*{\algorithmicindent}\hspace*{\algorithmicindent}$c \gets (T_{t-4}, \dots, T_{t-1}, T_t, \bar{T}_{\text{annual}})$
    \EndFor
    \State Store member trajectory:
    \Statex \hspace*{\algorithmicindent}\hspace*{\algorithmicindent}$\hat{\mathcal{T}} \gets \hat{\mathcal{T}} \cup \{\mathcal{T}^{(i)}\}$
\EndFor
\State \Return Full ensemble forecast $\hat{\mathcal{T}}$
\end{algorithmic}
\end{algorithm}

\subsection{Implementation and training}
\label{sec:implementation}

The vector field $v_\theta$ driving the flow matching process is parameterized using a PDE-Transformer architecture with a patch size of $4 \times 4$ grid cells \citep{holzschuh2025pde}. Network weights are optimized using the AdamW optimizer with an initial learning rate of $1 \times 10^{-4}$. The learning rate is dynamically adjusted using a cosine annealing schedule with a minimum learning rate ($\eta_{\text{min}}$) of $1 \times 10^{-5}$ \citep{loshchilov2016sgdr}. 

To ensure training stability and prevent overfitting, several regularization techniques are employed. We use a weight decay of $1 \times 10^{-7}$ and an early stopping mechanism with a patience of 50 epochs based on validation performance. Gradient explosion is mitigated using an Exponential Moving Average (EMA) gradient clipper (with decay coefficients $\beta_1 = 0.9$ and $\beta_2 = 0.99$, a maximum norm ratio of 2.0, and a clip norm ratio of 1.1) \citep{holzschuh2025pde}. Furthermore, mixed-precision training is used to speed-up training in the first 100 epochs.

We trained the model in three stages. First, we trained using a single member of EC-Earth3 (member 1), a batch size of 64 and a latitude weighted loss function on an NVDIA RTX 3090 GPU with 24 GB. After this initial training, we gained access to a GPU with more memory and so we increased the batch size to 704 (the largest that would fit on the GPU) and used the same EC-Earth3 member (member 1), and at the same time we moved to the MSE loss. Finally, we increased the number of EC-Earth3 members used for training to five (members 1 to 5). This multi-stage training strategy was adopted due to the availability of compute resources across the team. It is unclear if the multi-stage training was ultimately a benefit. Adjustments to the loss function and batch size may have played a role, although further experimentation is needed to confirm this, but the increase in the number of EC-Earth3 members available for training almost certainly improved the results. Training the final model took less than 28 hours in total.\\

\section{Results}\label{sec:results}

In this section we compare our EC-EarthFlow predictions with EC-Earth3. We focus on the northern hemisphere summer. Each prediction starts in May of a selected year between 1950 and 2150, and the only information taken from the corresponding EC-Earth3 member is the annual mean temperature and the first five days of daily temperature. From each such member we generate a 10-member EC-EarthFlow ensemble using 50 ODE steps, over both a short ($\sim$1-month) and a long ($\sim$8-month) rollout. Each EC-Earth3 member is compared with the EC-EarthFlow ensemble generated from it. Furthermore, EC-Earth3 members are also compared with one another, which indicates the spread expected between independent realisations. Sections \ref{sec:pointwise}, \ref{sec:ens}, \ref{sec:trend} and \ref{sec:dynamics} use a randomly chosen test member (member 16). Section \ref{sec:trend} additionally compares ensembles generated from members 15 and 16, and compares pairs of EC-Earth3 members (members 13--16).

\subsection{Pointwise predictions}\label{sec:pointwise}
EC-EarthFlow is successful at simulating daily temperatures at the grid point scale, including day-to-day variability, the annual cycle and the long term trend. Detailed analysis of the output for the gridpoints nearest to De Bilt in the Netherlands and to Sydney in Australia, highlights the model captures locally specific processes (Figure \ref{fig:points}). In EC-Earth3, Sydney experiences an increase in the daily temperature variability in the October-December period, which is reflected in the larger daily variability in EC-EarthFlow. Similarly, in De Bilt larger temperature variability in the extended winter season (from November) is well captured by EC-EarthFlow. It is remarkable that EC-EarthFlow can simulate such seasonal dependent increases in variability given only five days of temperatures at the end of April and the annual-mean temperature are used as inputs to the model. 

\subsection{Ensemble spread}\label{sec:ens}
\begin{figure*}[t]
 \centerline{\includegraphics[width=0.99\textwidth]{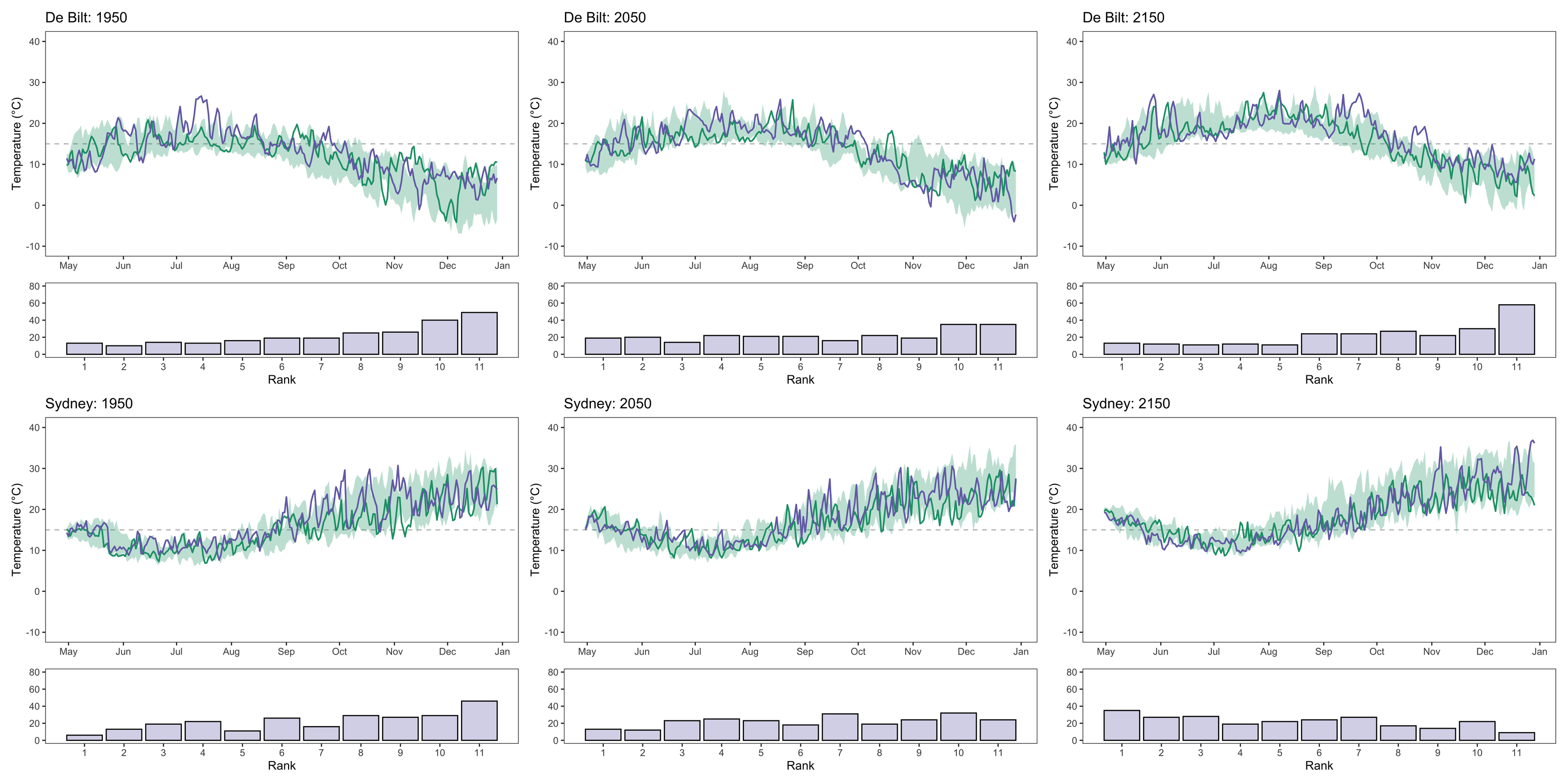}}
  \caption{Daily temperature at two locations (De Bilt, The Netherlands (top), and Sydney, Australia (bottom)) from May - November in 1950 (left), 2050 (middle) and 2150 (right). Predictions are made using a rollout length of 245 steps from only 5 EC-Earth member states at the end of April (member 16). For comparison the full EC-Earth simulation is shown in purple. The green shading indicates the ensemble range from nine EC-EarthFlow members, and the green line shows predictions from a single random EC-EarthFlow member. The gray line indicates a daily temperature value of 15$^{\circ}$C as reference. The histograms show counts of the rank that the EC-Earth3 member is in the 10-member EC-EarthFlow ensemble.}
  \label{fig:points}
\end{figure*}

\begin{figure*}[t]
 \centerline{\includegraphics[width=0.99\textwidth]{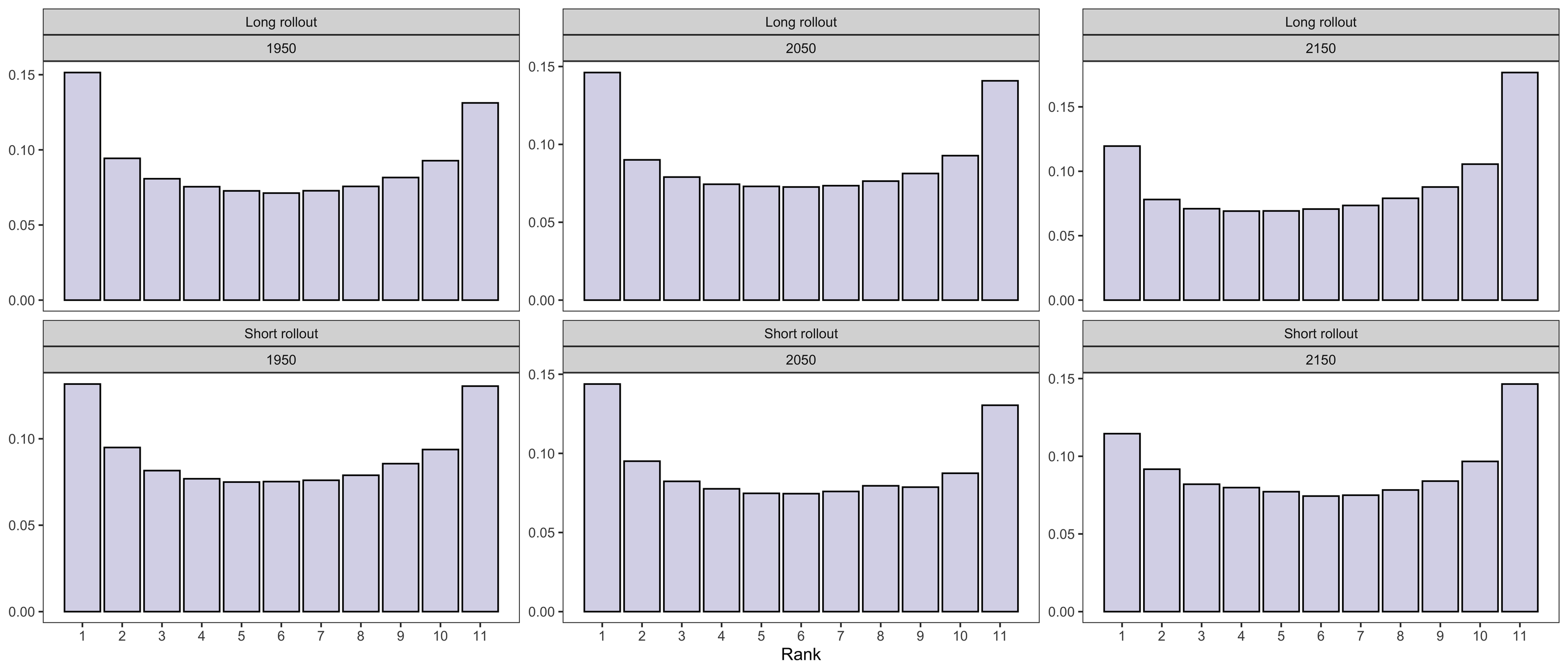}}
  \caption{Rank histograms of all grid points in 1950 (left), 2050 (middle) and 2150 (right). EC-EarthFlow predictions are made using the long (top) and short (bottom) rollout length from May in each year and inputs from a single EC-Earth member (member 16).}
  \label{fig:rh_allpoints}
\end{figure*}

We use rank histograms to more formally assess the realism of the ensemble spread, which compares the rank of the EC-Earth3 member to the EC-EarthFlow ensemble. The histogram should be flat if EC-Earth3 is statistically indistinguishable from the EC-EarthFlow ensemble. The signal is mixed across different locations and years, as expected given the relatively short time periods used to calculate the rank histogram. EC-Earth3 has warmer temperatures (EC-EarthFlow has cooler temperatures), especially in De Bilt in 1950 and 2150 (Figure \ref{fig:points}a,e), while opposite patterns can be seen in other locations and years. The rank histograms over all grid points shows that the EC-EarthFlow ensemble tends to be under-dispersed, as the EC-Earth3 member is too often above the highest EC-EarthFlow member or below the lowest EC-EarthFlow member (Figure \ref{fig:rh_allpoints}).

\subsection{Long-term trends}\label{sec:trend}

\begin{figure*}[t]
  \centerline{\includegraphics[width=0.99\textwidth]{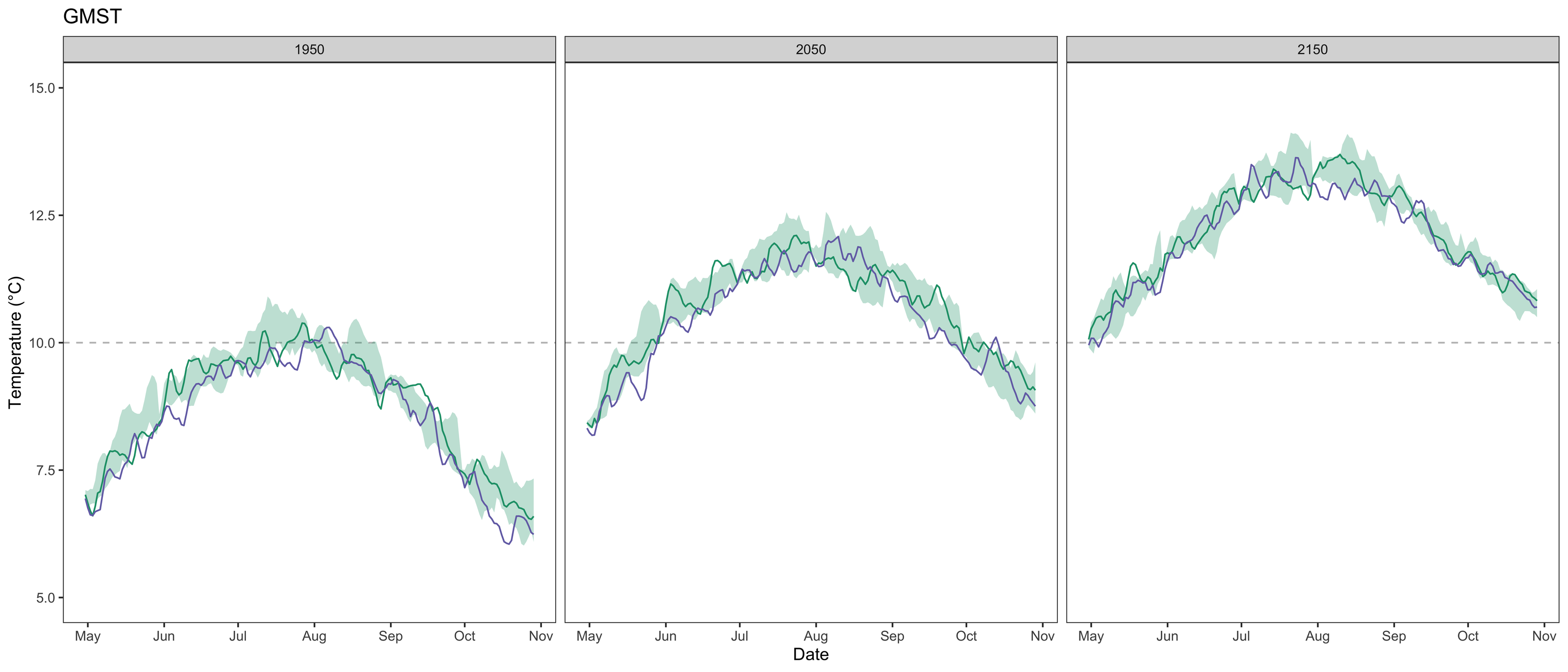}}
  \caption{Global mean surface temperature with short rollout, starting in each month of the extended northern Hemisphere summer period (May - October). Each panel shows projections from a different year (1950, 2050, 2150). Inputs are from a single test member of EC-Earth3 (member 16, purple line). The green shading indicates the ensemble range from nine EC-EarthFlow members, and the green line shows predictions from a single random EC-EarthFlow member. The gray line indicates a GMST of 10$^{\circ}$C as reference.}\label{fig:gmst_sim16}
\end{figure*}

Next, we show that EC-EarthFlow is able to reproduce both the annual cycle and long-term trend in global mean surface temperature (GMST, Figure \ref{fig:gmst_sim16}). The annual cycle in GMST typically peaks in the northern hemisphere summer driven by the larger land area in this hemisphere. This is well simulated by EC-Earth3 and captured by the EC-EarthFlow emulator, which makes sense given that the annual mean temperature from EC-Earth3 is used as an input to the model. The magnitude of the forced response to external forcing is also well captured by EC-EarthFlow compared to EC-Earth3, with northern hemisphere summer temperatures around 2.5$^{\circ}$C warmer by 2150 (Figure \ref{fig:gmst_sim16}).\\

\begin{figure*}[t]
 \centerline{\includegraphics[width=0.99\textwidth]{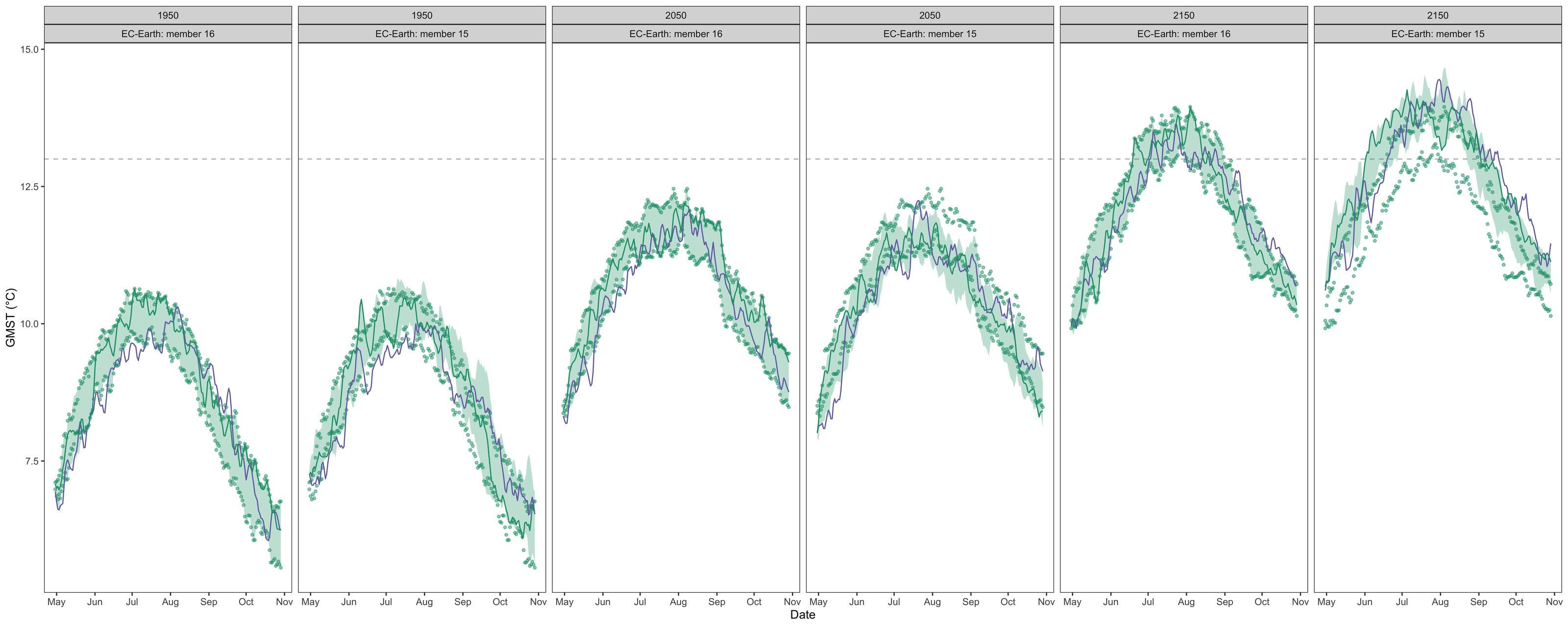}}
  \caption{Global mean surface temperature (long rollout) (May - November). Columns show projections in select years (1950, 2050, 2150) made using EC-Earth3 members 16 and 15 as inputs (purple lines). The green shading indicates the ensemble range from all EC-EarthFlow members, and the green line shows predictions from a single random EC-EarthFlow member. The green dots show the EC-EarthFlow ensemble range from projections generated from EC-Earth3 member 16.  The gray line indicates a GMST of 13$^{\circ}$C as reference.}\label{fig:gmst_membs}
\end{figure*}

Besides the annual cycle and long-term trend, the EC-Earth3 ensemble contains inter-annual GMST variability due to natural climate variability (e.g. El-Ni\~no Southern Oscillation, ENSO). For example, EC-Earth3 member 16 is warmer than member 15 in 2150 (Figure \ref{fig:gmst_membs}). This variability is generally preserved in the EC-EarthFlow ensemble as the generated ensemble tends to follow the warming pattern of the EC-Earth3 member used as conditioning input. This is expected behavior given the inclusion of annual mean temperature as a conditioning variable. It is promising to see that, despite the heavy influence of the EC-Earth test member, there is spread between the EC-EarthFlow members, with overlap between EC-EarthFlow ensemble members generated from EC-Earth3 member 15 and 16 (Figure \ref{fig:gmst_membs}). \\

\begin{figure*}[t]
 \centerline{\includegraphics[width=0.99\textwidth]{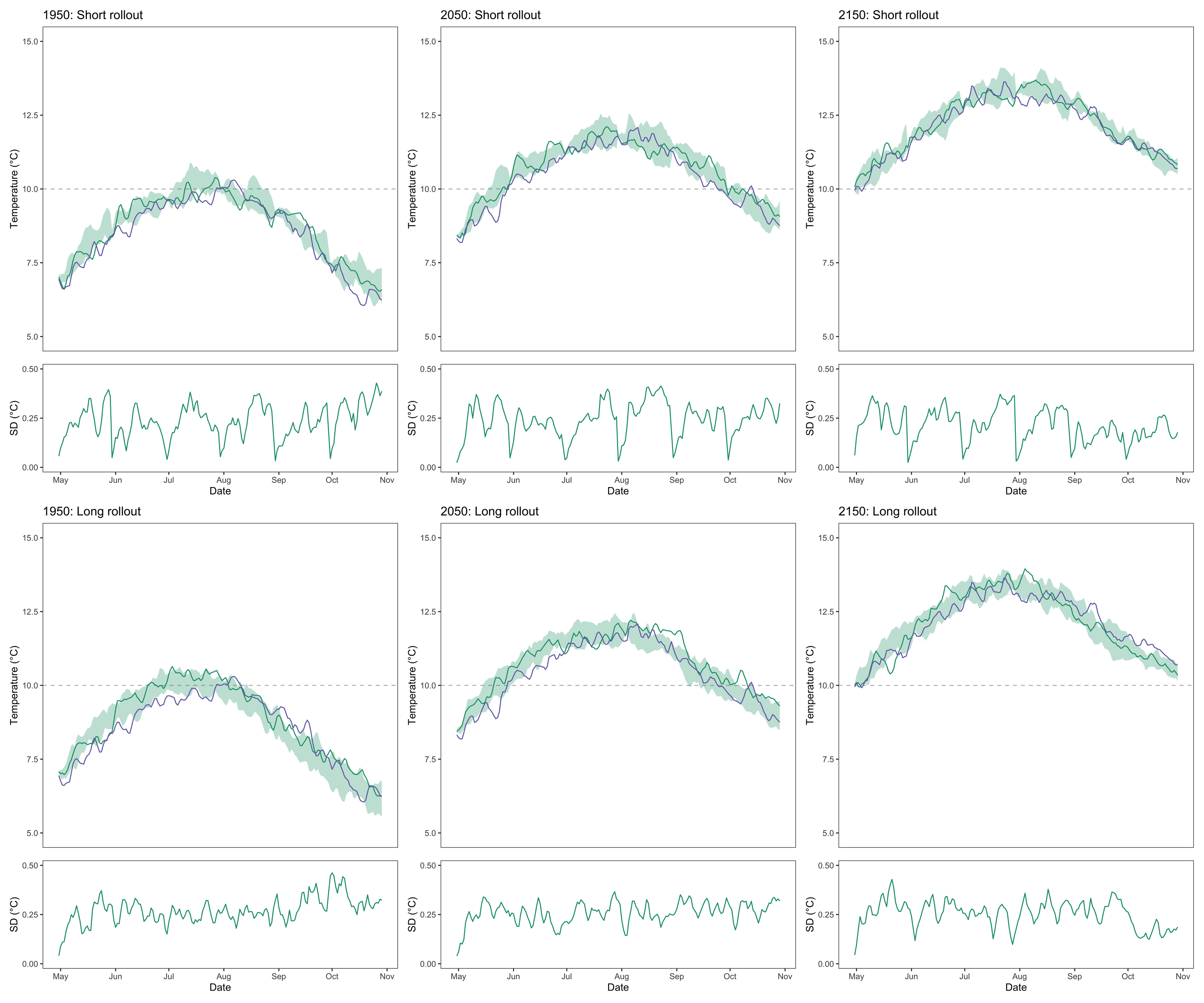}}
  \caption{Global mean surface temperature projections from May - November with a short (top row, i.e., consecutive 30-day rollouts) and long (bottom row, i.e., a single rollout from May to November) lengths in 1950, 2050, and 2150. The black line is the `truth' from EC-Earth3. The green lines are ensemble members created by the flow matching model. The red line indicates a GMST of 10$^{\circ}$C as reference.}\label{fig:gmst_rollout}
\end{figure*}

We demonstrate that predictions from EC-EarthFlow are stable for long rollout periods by comparing short and long rollouts (Figure \ref{fig:gmst_rollout}). In the short rollout length (30 days), we concatenate 7 individual monthly predictions together to make a single extended season. This ignores the temporal dependencies between EC-EarthFlow ensemble members but is done here to illustrate that it is possible to use rollouts of varying lengths. It is evident in the short rollout length that the trajectory is brought back towards the EC-Earth3 test member at the start of each inference period, e.g., at the start of June in 1950 (Figure \ref{fig:gmst_rollout}). This is also evident in the daily standard deviation of the EC-EarthFlow ensemble in short rollout lengths. Despite this un-desirable feature, the remainder of the trajectory is well simulated with comparable spread between the EC-EarthFlow members in the long and short rollouts. The spread of the long rollouts increases over the first few time steps and is then consistent throughout the extended season (Figure \ref{fig:gmst_rollout}). The annual cycle and trends are similar in both rollouts despite the fact that the long rollout has limited influence from the daily temperature variability in the EC-Earth3 test member, and the short rollout has daily temperatures from the EC-Earth3 test member injected throughout the extended season. This shows that the model is stable for long periods and that errors are not accumulating to produce instabilities.

\begin{figure*}[t]
 \centerline{\includegraphics[width=0.99\textwidth]{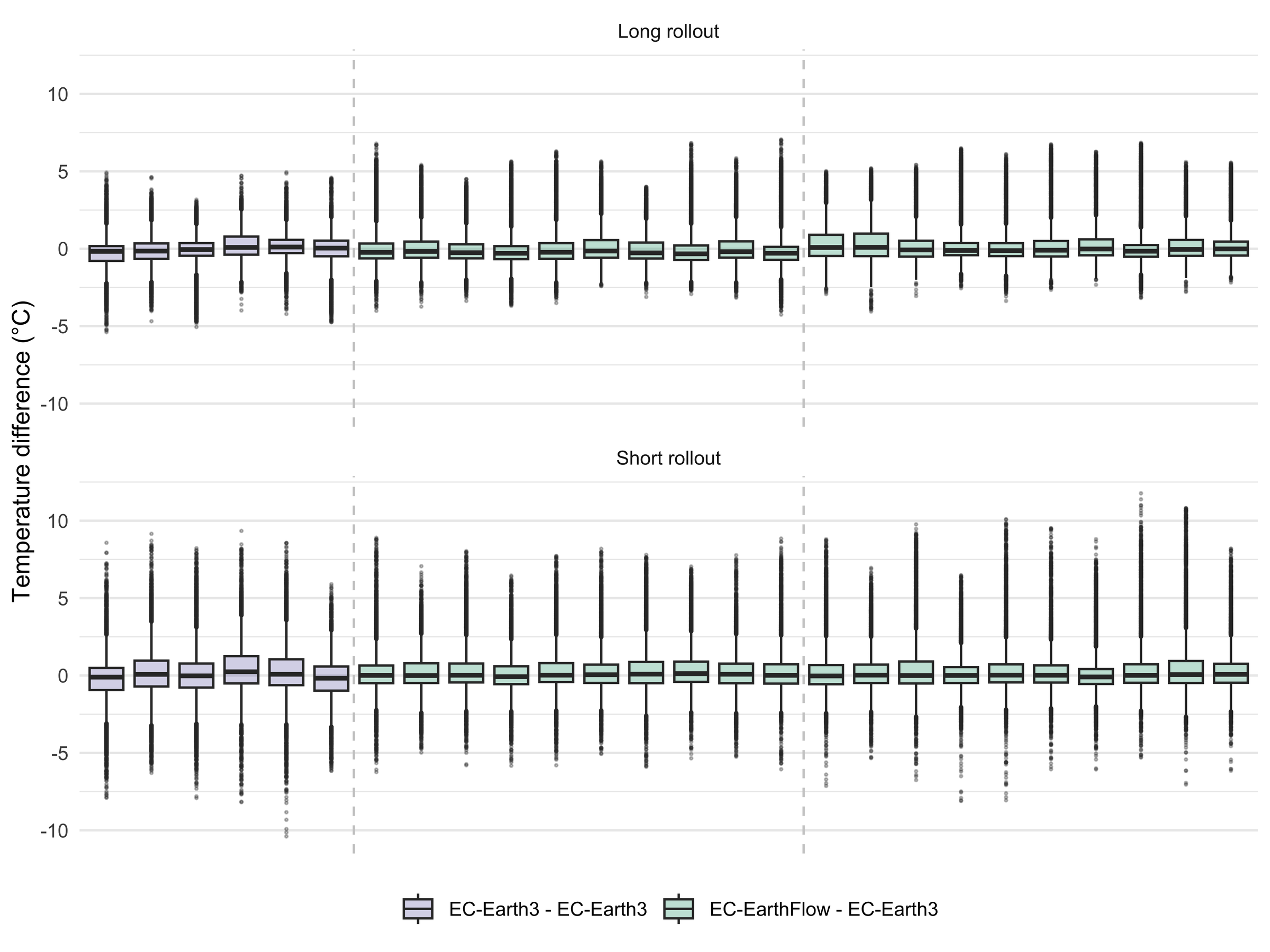}}
  \caption{Box-plots of the grid point wise trajectory mean temperature differences between EC-Earth3 members (purple) and between predictions from EC-EarthFlow and the EC-Earth3 member used to generate them (green) in 2050. Similar results are seen for other years and other EC-EarthFlow members.}\label{fig:mean_bias}
\end{figure*}

We show the mean temperature differences between EC-Earth members and between EC-Earth and EC-EarthFlow to assess how errors accumulate over the rollout period and to put these in the context of expected differences between different members of a GCM. We expect differences between these two predictions, given that we are looking at climate simulations rather than weather forecasts, which should sample the range of natural variability. We expect however that these differences are somewhat random, and reflect differences in the location of weather patterns, rather than being systematic. Indeed, this is what we see when looking at the EC-Earth simulations, as the largest differences are generally evenly distributed around zero (the dots on the purple boxes in Figure \ref{fig:mean_bias}). Overall the mean differences between EC-Earth and EC-EarthFlow are consistent with what we would expect. There are some deficiencies that can be seen here. First, the differences between the generated and EC-Earth members are smaller than the differences between two EC-Earth members. This is consistent with the too small ensemble spread that we saw in Figure \ref{fig:rh_allpoints}. Furthermore, we see a bias in the EC-EarthFlow members as there are more grid points with positive bias. This stems from a warm bias at the poles in the winter hemisphere. Finally, we see that the longer rollout accumulates these issues more than the shorter rollout (Figure \ref{fig:mean_bias}). 

\subsection{Large-scale dynamics}\label{sec:dynamics}

\begin{figure*}[t]
 \centerline{\includegraphics[width=0.99\textwidth]{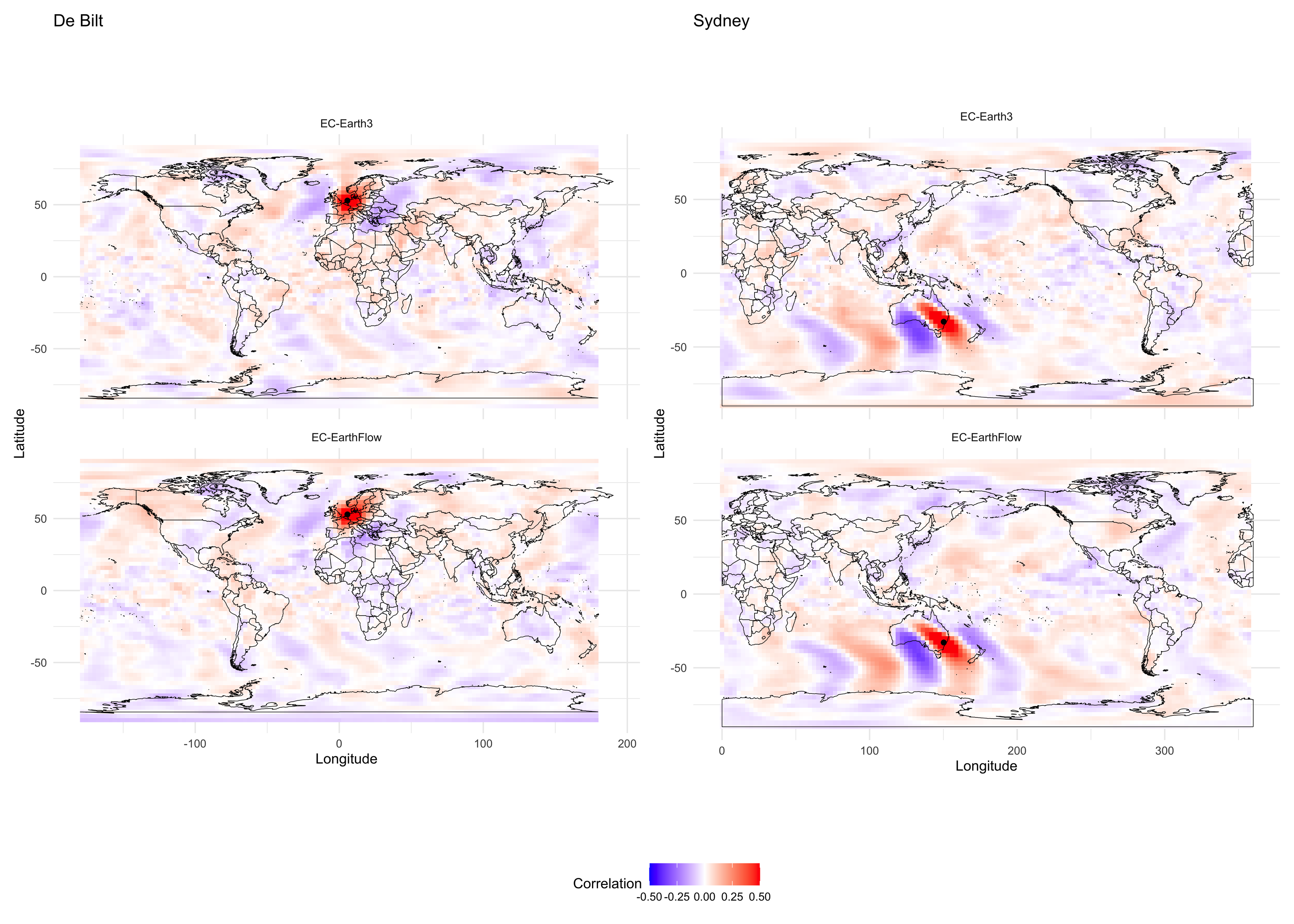}}
  \caption{The correlation between temperature in De Bilt (left) or Sydney (right) and the rest of the world in EC-Earth3 (top) and EC-EarthFlow (bottom). We use a high-pass filter to remove variability and then take the correlation over all days.}\label{fig:correl}
\end{figure*}

Finally, we show that EC-EarthFlow has successfully learned spatial dependencies due to large-scale dynamical processes in the atmosphere as simulated in EC-Earth3. We remove high-frequency variability in daily temperature then take the correlation (over all inference days) between two central points and all other points. EC-EarthFlow is able to reproduce the spatial auto-correlation and teleconnections of temperature as simulated by EC-Earth3. It is clear that hot temperatures over western Europe (e.g., in De Bilt, the Netherlands) are associated with cooler regions up and downstream. This is consistent with atmospheric blocking conditions, often in atmospheric wave trains, and advective and radiative processes leading to cold and warm temperatures (Figure \ref{fig:correl}). This dynamical pathway is simulated explicitly in EC-Earth3, but is learnt only from 2m-temperature in the EC-EarthFlow emulator. This suggests that the emulator can learn multivariate physical processes at play in the atmosphere from only a single variable. The strongest correlation in the wave patterns appear to be reproduced well by EC-EarthFlow, although there is some smoothing compared to EC-Earth3 in regions further from the point of interest, for example, in the North Pacific ocean in the Sydney-centered plots. While this might be an artifact in EC-Earth3, we would still hope that the flow matching model is able to reproduce the training data perfectly (Figure \ref{fig:correl}). 

\section{Discussion and Conclusions}\label{sec:conc}

These results demonstrate that it is possible to emulate the non-stationary daily climate variability of EC-Earth3 using a generative flow matching model, EC-EarthFlow, with only a single variable as an input. We successfully emulate the annual cycle, long-term warming trend, and physical relationships as simulated by the global climate model. We showed that the emulator is stable for long rollout periods of up to an extended season. Future work should explore the stability and quality of the emulated trajectories for annual and multi-annual rollout lengths. 

At the local scale, the emulator is able to learn not only the increasing warming trend between 1950 and 2150, but also the increased variability of temperature in eastern Australia in the southern hemisphere spring-summer period. This increased variability is possibly related to ENSO, and it is remarkable that the emulator learns it and is able to simulate it in the long rollout period. The only context information that the flow matching model has received from EC-Earth3 for these inference periods is daily temperatures from late April (the five days before the first predicted days shown in Figure \ref{fig:points}) and the annual mean temperature. The spread of the EC-EarthFlow ensemble is generally good, if a little under-dispersed, but it could be further improved by tuning parameters in the flow matching model during training.

At the global scale, the emulator accurately reproduces the amplitude and phase of the GMST annual cycle, including the stronger signal during Northern Hemisphere summer driven by land–sea contrast. Again, these features are well simulated with the long rollout period with limited context input from the GCM. Importantly, the emulator also learns the global warming trend, indicating that it can represent the response to external forcing scenarios. This is a key requirement for climate model emulation.

The analysis of ensemble behavior highlights an important characteristic of the model: while the generated ensemble is strongly conditioned on the input EC-Earth3 member, likely due to the inclusions of the annual mean temperature, it still exhibits internal spread between generated members. This indicates that the emulator preserves part of the input state dependence while introducing stochastic variability. However, the dependence on the conditioning input may limit the ability of the model to fully explore the space of internal variability independently of the driving member. One possible solution is to source the conditioning trend input from other datasets during training and/or inference. Instead of using annual mean temperature from the same member, future work could explore the use of greenhouse gasses directly, the output of a reduced-complexity climate models (also known as "climate emulators") that replicate some aspects of a GCM, or statistical or machine learning emulators that predict annual temperatures from forcing information \citep[e.g.][]{kaltenborn2023climateset, nayak2025evaluating}. In this way, a wide variety of climate projections could be easily incorporated to this generative model which is then able to return an ensemble of daily temperatures for the range of warning scenarios.

The comparison of short and long rollouts shows that the model is stable for extended inference periods. This is a key challenge for generative models but demonstrates that the conditioning is sufficient to keep the model within reasonable bounds. The model struggled with temperatures at the poles, producing spatially coherent biases in these regions. The effect was most evident at the pole in the winter season (i.e., Antarctica in northern hemisphere summer), suggesting that it was possibly related to the larger variability of the temperature. We explored different normalization methods, including normalization per grid point, but saw little improvement. This suggest that the effect is likely due to the use of a regular lat-lon grid, rather than a more complex projection that better handles the varying grid sizes at the poles. Future work can explore the use of different projections, such as the HEALPix grid used in recent studies \citep[e.g.][]{brenowitz2025climate}.

Finally, the spatial correlation analysis provides strong evidence that EC-EarthFlow has learned aspects of the underlying atmospheric dynamics. The reproduction of teleconnection-like wave patterns and blocking signatures—despite being trained only on 2m temperature—suggests that the model implicitly captures large-scale circulation constraints. The smoother correlations at large distances indicate some loss of higher-order structure, which may reflect limitations in model capacity or training objectives. Overall these results can give users confidence that the model is learning the physics of the atmosphere.

Results from this proof-of-concept paper are promising, but there is a long list of potential improvements or extensions, apart from those listed above. It is encouraging from a computational perspective that we show the whole atmosphere is not needed to simulate key features of climate variability and trend with long rollout periods. While 2m-temperature is an important variable, it is clearly not the only variable of interest to scientists assessing future (regional) climate changes. As such, the inclusion of additional variables, like precipitation, wind speed, humidity or soil moisture, is an obvious next step. Additionally, higher spatial resolution is required for this to be of real use. Emulation of new km-scale GCMs is an important research topic for the coming years. 

Beyond adding variables and increasing resolution, several improvements concern the generative model itself. One could replace flow matching with alternatives such as stochastic interpolants \citep{albergo2025stochastic, chen2024probabilistic}, which model the transition between states directly through an SDE rather than solving an ODE from a Gaussian sample. A related option is diffusion, which is equivalent to flow matching up to a reparameterization, so that a model trained with flow matching can be sampled with an SDE without retraining. Knowledge of the underlying physics can also be incorporated, either during training or at inference time \citep{mucke2026physics, debeire2026physics}. Inference cost can be reduced by using consistency or flow map models, which replace the expensive ODE solve with a few model evaluations \citep{boffi2026build}. After deployment, predictions can be improved by assimilating available observations to limit drift from the true state. This can be done with the current flow matching model directly via guidance \citep{amoros2026guiding} and can be combined with conventional data assimilation so that techniques such as localization carry over \citep{seabra2026integrating}. Finally, the PDE transformer architecture may benefit from better conditioning on the history, which could in turn allow conditioning on longer histories.

\acknowledgments
The authors thank Benjamin Sanderse, Mohit Anand, Bastien Francois and Jasper Wijnands for helpful advice. We thank the Dutch community for AI in weather and climate: AIMET-NL (www.aimet.nl). This work used the Dutch national e-infrastructure with the support of the SURF Cooperative using grant no. EINF-10742.

\datastatement

The specific EC-Earth3 simulations used to train and evaluate the emulator are held at KNMI. They are available on request from K. Whan (kirien.whan@knmi.nl), Royal Netherlands Meteorological Institute. Documentation of the simulation setup and of all pre-processing applied before training is provided in Section 2. Code for the emulator is available at https://github.com/nmucke/scientific-stochastic-interpolants.

\bibliographystyle{ametsocV6}
\bibliography{references}

\end{document}